\documentclass[letterpaper]{article}

\usepackage{aaai}
	
\usepackage{times}
\usepackage{helvet}
\usepackage{courier}
\usepackage{latexsym}

\usepackage{amsfonts,eucal,amsbsy,amsopn}
\usepackage{amsmath,amssymb,amsthm,stmaryrd,color}
\usepackage{graphicx}
\usepackage{multirow,rotating}
\usepackage{xspace}
\usepackage{dsfont}
\usepackage{url}
\usepackage{verbatim}
\usepackage[font=small]{caption}
\usepackage{mdframed}

\usepackage[tight]{subfigure}
\usepackage{subfloat}
\usepackage{framed}
\usepackage{mdframed}
\usepackage{flushend}

\usepackage{tabularx}
\usepackage{booktabs}

\newcommand{\myciteauthoryear}[1]{\citeauthor{#1} (\citeyear{#1})}

\newcommand{\ignore}[1]{}

\def\eqnref#1{Eqn.~\ref{#1}}

\begin{document}

\title{Listen, Attend, and Walk: Neural Mapping\\ of Navigational Instructions to Action Sequences}

\author{
  Hongyuan Mei \qquad Mohit Bansal \qquad Matthew R.~Walter\\
  Toyota Technological Institute at Chicago\\
  Chicago, IL 60637 \\
  \{\texttt{hongyuan,mbansal,mwalter\}@ttic.edu}}

\maketitle

\begin{abstract}
    We propose a neural sequence-to-sequence model for direction
    following, a task that is essential to realizing effective
    autonomous agents. Our alignment-based encoder-decoder model with
    long short-term memory recurrent neural networks (LSTM-RNN)
    translates natural language instructions to action sequences based
    upon a representation of the observable world state. We
    introduce a multi-level aligner that empowers our model to focus
    on sentence ``regions'' salient to the current world state by
    using multiple abstractions of the input sentence. In contrast to
    existing methods, our model uses no specialized linguistic
    resources (e.g., parsers) or task-specific annotations (e.g., seed
    lexicons). It is therefore generalizable, yet still achieves
    the best results reported to-date on a benchmark single-sentence
    dataset and competitive results for the limited-training
    multi-sentence setting.  We analyze our model through a series of
    ablations that elucidate the contributions of the primary
    components of our model.
\end{abstract}
\section{Introduction}
\label{sec:introduction}

\begin{figure}[!tb]
    \centering
    \includegraphics[width=1.0\linewidth]{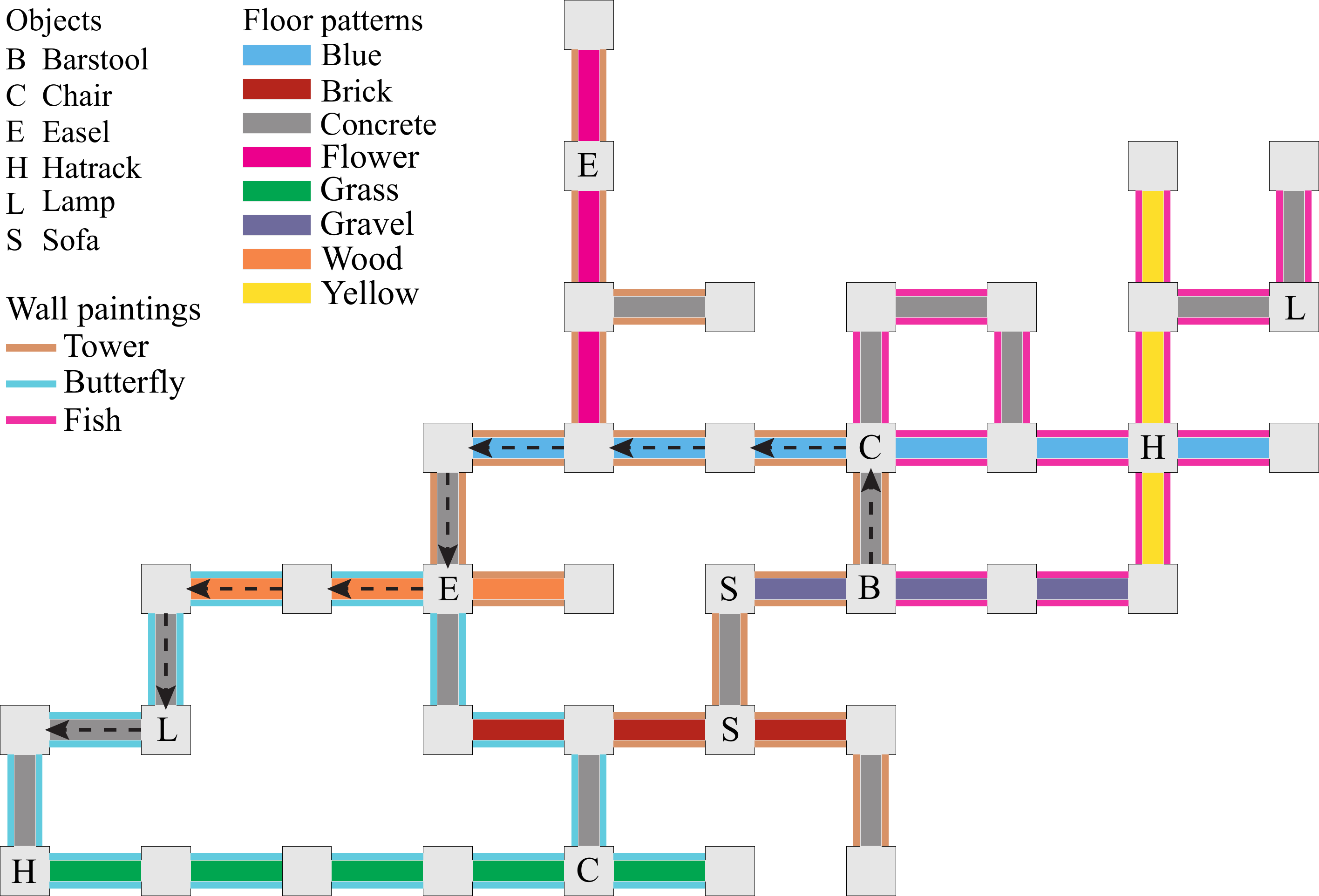}
    %
    \begin{mdframed}[align=center]
    {\footnotesize
      Place your back against the wall of the ``T'' intersection.
      Go forward one segment to the intersection with the blue-tiled hall.
      This interesction [sic] contains a chair.
      Turn left.
      Go forward to the end of the hall.
      Turn left.
      Go forward one segment to the intersection with the wooden-floored hall.
      This intersection conatains [sic] an easel.
      Turn right. 
      Go forward two segments to the end of the hall. 
      Turn left.
      Go forward one segment to the intersection containing the lamp.
      Turn right. 
      Go forward one segment to the empty corner. 
    }
    \end{mdframed}
    \vspace{7.5pt}
    \caption{An example of a route instruction-path pair in one of the
      virtual worlds from~\myciteauthoryear{macmahon-06} with colors
      that indicate floor patterns and wall paintings, and letters
      that indicate different objects. Our method successfully infers
      the correct path for this instruction.} \label{fig:mapexample}
\end{figure}

Robots must be able to understand and successfully execute natural
language navigational instructions if they are to work seamlessly
alongside people. For example, someone using a voice-commandable
wheelchair might direct it to ``Take me to the room across from the
kitchen,'' or a soldier may command a micro aerial vehicle to ``Fly
down the hallway into the second room on the right.'' However,
interpreting such free-form instructions (especially in unknown
environments) is challenging due to their ambiguity and complexity,
such as uncertainty in their interpretation (e.g., which hallway does
the instruction refer to), long-term dependencies among both the
instructions and the actions, differences in the amount of detail
given, and the diverse ways in which the language can be
composed. Figure~\ref{fig:mapexample} presents an example instruction
that our method successfully follows.

Previous work in this domain
\cite{chen-11,chen-12,kim-12,kim-13,artzi-13,artzi-14} largely
requires specialized resources like semantic parsers, seed lexicons,
and re-rankers to interpret ambiguous, free-form natural language
instructions. In contrast, the goal of our work is to learn to map
instructions to actions in an end-to-end fashion that assumes no prior
linguistic knowledge. Instead, our model learns the meaning of all the
words, spatial relations, syntax, and compositional semantics from
just the raw training sequence pairs, and learns to to translate the
free-form instructions to an executable action sequence.

We propose a recurrent neural network with long short-term memory
(LSTM) \cite{hochreiter-97} to both encode the navigational
instruction sequence bidirectionally and to decode the representation
to an action sequence, based on a representation of the current world
state. LSTMs are well-suited to this task, as they have been shown to
be effective in learning the temporal dependencies that exist over
such sequences, especially for the tasks of image captioning, machine
translation, and natural language generation
~\cite{kiros-14,donahue-14,chen-15,karpathy-15,vinyals-14,sutskever-14,rush-15,wen-15}. Additionally,
we learn the correspondences between words in the input navigational
instruction and actions in the output sequence using an
alignment-based LSTM~\cite{bahdanau-14,xu-15}. Standard alignment
methods only consider high-level abstractions of the input (language),
which sacrifices information important to identifying these
correspondences. Instead, we introduce a \emph{multi-level} aligner
that empowers the model to use both high- and low-level input
representations and, in turn, improves the accuracy of the inferred
directions.

We evaluate our model on a benchmark navigation
dataset~\cite{macmahon-06} and achieve the best results reported
to-date on the single-sentence task (which contains only 2000 training
pairs), without using any additional resources such as semantic
parsers, seed lexicons, or rerankers used in previous work.  On the
multi-sentence task of executing a full paragraph, where the amount of
training pairs is even smaller (just a few hundred pairs), our model
performs better than several existing methods and is competitive with
the state-of-the-art, all of which use specialized linguistic
resources, extra annotation, or reranking. We perform a series of
ablation studies in order to analyze the primary components of our
model, including the encoder, multi-level representations, alignment,
and bidirectionality.

\section{Related Work} \label{sec:related}

A great deal of attention has been paid of late to algorithms that
allow robots and autonomous agents to follow free-form navigational
route
instructions~\cite{macmahon-06,kollar-10,chen-11,chen-12,kim-12,kim-13,kong-14,hemachandra-15}.
These methods solve what~\myciteauthoryear{harnad-90} refers to as the
symbol grounding problem, that of associating linguistic elements with
their corresponding manifestation in the external world. Initial
research in natural language symbol grounding focused on
manually-prescribed mappings between language and sets of predefined
environment features and actions~\cite{winograd-70,macmahon-06}. More
recent work in statistical language understanding learns to convert
free-form instructions into their referent symbols by observing the
use of language in a perceptual context~\cite{mooney-08}. These
methods represent natural language grounding in terms of manually
defined linguistic, spatial, and semantic
features~\cite{kollar-10,matuszek-10,tellex-11}. They learn the model
parameters from natural language corpora, often requiring expensive
annotation to pair each phrase to its corresponding grounding.
\begin{figure*}[!t]
    \centering
    \includegraphics[width=0.8\linewidth]{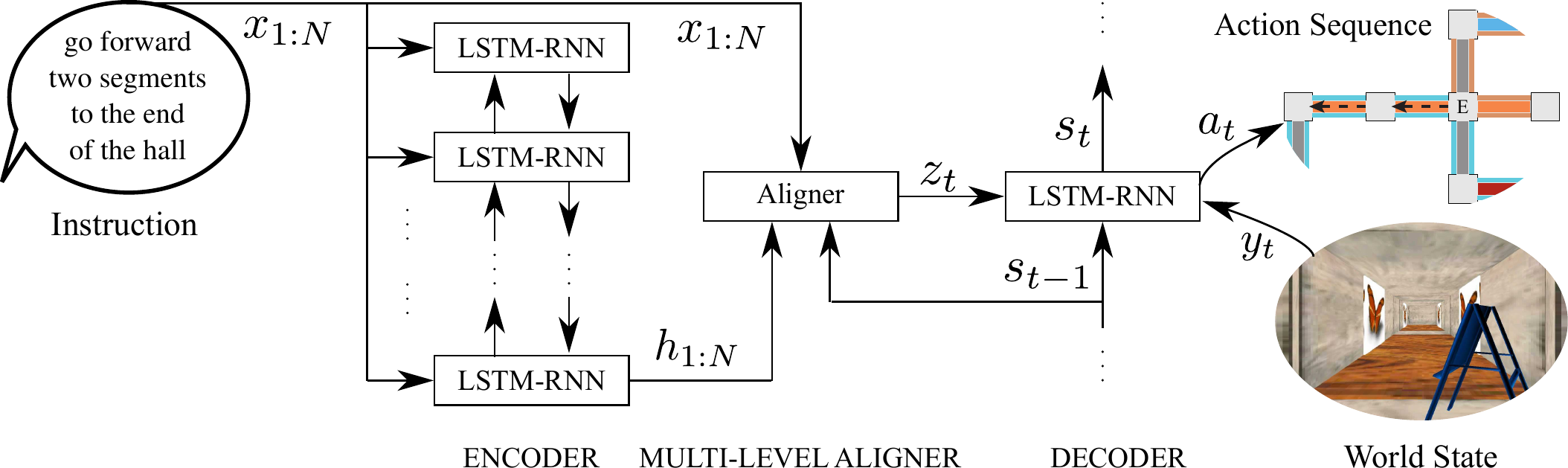}
    \caption{Our encoder-aligner-decoder model with multi-level
      alignment} \label{fig:model}
\end{figure*}

One class of grounded language acquisition methods treats the language
understanding problem as one of learning a parser that maps free-form
language into its formal language equivalent. For example,
\myciteauthoryear{matuszek-10} assume no prior linguistic knowledge
and employ a general-purpose supervised semantic parser
learner. Alternatively, \myciteauthoryear{chen-11} parse free-form
route instructions into formal action specifications that a robot
control process can then execute. They learn the parser in a weakly
supervised manner from natural language instruction and action
sequence pairs, together with the corresponding world
representation. Alternatively, \myciteauthoryear{kim-12} frame
grounded language learning as probabilistic context free grammar
(PCFG) induction and use learned lexicons~\cite{chen-11} to control
the space of production rules, which allows them to scale PCFGs to the
navigation domain. \myciteauthoryear{kim-13} improve upon the accuracy
by adding a subsequent re-ranking step that uses a weakly supervised
discriminative classifier. Meanwhile, \myciteauthoryear{artzi-13}
learn a combinatory categorical grammar (CCG)-based semantic parser to
convert free-form navigational instructions to their manifestation in
a lambda-calculus representation. As with \myciteauthoryear{kim-13},
they improve upon the accuracy through
re-ranking. \myciteauthoryear{artzi-14} extend their CCG parser
learning by using statistics of the corpus to control the size of the
lexicon, resulting in improved multi-sentence accuracy.

A second class of grounded language learning techniques function by
mapping free-form utterances to their corresponding object, location,
and action referents in the agent's world model. These methods learn a
probabilistic model that expresses the association between each word
in the instruction and its matching referent in the world model. The
problem of interpreting a new instruction then becomes one of
inference in this learned model. \myciteauthoryear{kollar-10} build a
generative model over the assumed flat, sequential structure of
language that includes a combination of pre-specified and learned
models for spatial relations, adverbs, and
verbs. \myciteauthoryear{tellex-11} later propose a discriminative
model that expresses the hierarchical, compositional structure of
language. They factor the probability distribution according to the
parse structure of the free-form command and employ a log-linear
factor graph to express the learned correspondence between linguistic
elements and the space of groundings (objects, locations, and
actions).

We adopt an alternative formulation and treat the problem of
interpreting route instructions as a sequence-to-sequence learning
problem. We learn this mapping in an end-to-end fashion using a neural
network, without using any prior linguistic structure, resources, or
annotation, which improves generalizability.  Our method is inspired
by the recent success of such sequence-to-sequence methods for machine
translation~\cite{sutskever-14,bahdanau-14,cho14}, image and video
caption
synthesis~\cite{kiros-14,mao-14,donahue-14,vinyals-14,chen-15,karpathy-15},
and natural language generation~\cite{rush-15,wen-15}, which similarly
adopt an encoder-decoder approach. Our model encodes the input
free-form route instruction and then decodes the embedding to identify
the corresponding output action sequence based upon the local,
observable world state (which we treat as a third sequence type that
we add as an extra connection to every decoder step). Moreover, our
decoder also includes alignment to focus on the portions of the
sentence relevant to the current action, a technique that has proven
effective in machine translation~\cite{bahdanau-14} and machine
vision~\cite{mnih-14,ba-14,xu-15}. However, unlike the standard
alignment techniques, our model learns to align based not only on the
high-level input abstraction, but also the low-level representation of
the input instruction, which improves performance. Recently,
\myciteauthoryear{andreas-15} use a conditional random field model to
learn alignment between instructions and actions; our LSTM-based
aligner performs substantially better than this approach.

\section{Task Definition} \label{sec:taskdefinition}

We consider the problem of mapping natural language navigational
instructions to action sequences based only on knowledge of the local,
observable environment. These instructions may take the form of
isolated sentences (single-sentence) or full paragraphs
(multi-sentence). We are interested in learning this mapping from
corpora of training data of the form $(x^{(i)}, a^{(i)}, y^{(i)})$ for
$i = 1,2,\ldots,n$, where $x^{(i)}$ is a variable length natural
language instruction, $a^{(i)}$ is the corresponding action sequence,
and $y^{(i)}$ is the observable environment representation. The model
learns to produce the correct action sequence $a^{(i)}$ given a
previously unseen $(x^{(i)}, y^{(i)})$ pair. The challenges arise from
the fact that the instructions are free-form and complex, contain
numerous spelling and grammatical errors, and are ambiguous in their
meaning. Further, the model is only aware of the local environment in
the agent's line-of-sight.

In this paper, we consider the route instruction dataset generated by
\myciteauthoryear{macmahon-06}. The data includes free-form route
instructions and their corresponding action sequences within three
different virtual worlds. The environments (Fig.~\ref{fig:mapexample})
consist of interconnected hallways with a pattern (grass, brick, wood,
gravel, blue, flower, or yellow octagons) on each hallway floor, a
painting (butterfly, fish, or Eiffel Tower) on the walls, and objects
(hat rack, lamp, chair, sofa, barstool, and easel) at
intersections. After having explored an environment, instructors were
asked to give written commands that describe how to navigate from one
location to another without subsequent access to the map. Each
instruction was then given to several human followers who were tasked
with navigating in the virtual world without a map, and their paths
were recorded. Many of the raw instructions include spelling and
grammatical errors, others are incorrect (e.g., misusing ``left'' and
``right''), approximately $10\%$ of the single sentences have no
associated action, and $20\%$ have no feasible path.


\section{The Model}
\label{sec:model}

We formulate the problem of interpreting natural language route
instructions as inference over a probabilistic model
$P(a_{1:T} \vert y_{1:T}, x_{1:N})$, where
$a_{1:T} = (a_1,a_2,\ldots,a_{T})$ is the action sequence, $y_t$ is the
world state at time $t$, and $x_{1:N} = (x_1, x_2, \ldots, x_N)$ is the
natural language instruction,
\begin{subequations} \label{eqn:argmax}
\begin{align}
    a_{1:T}^* &= \underset{a_{1:T}}{\textrm{arg max }} P(a_{1:T}|y_{1:T},
    x_{1:N})\\
    &= \underset{a_{1:T}}{\textrm{arg max }} \prod\limits_{t=1}^T
    P(a_t \vert a_{1:t-1}, y_t, x_{1:N})
\end{align}
\end{subequations}

This problem can be viewed as one of mapping the given instruction sequence
$x_{1:N}$ to the action sequence $a_{1:T}$.  An effective means of learning
this sequence-to-sequence mapping is to use a neural encoder-decoder
architecture.  We first use a bidirectional recurrent neural network model
to encode the input sentence
\begin{subequations}
   \begin{align}
     h_j &= f(x_j, h_{j-1}, h_{j+1})\\
     z_t &= c(h_1, h_2, \ldots h_N),
   \end{align}
\end{subequations}
where $h_j$ is the encoder hidden state for word
\mbox{$j \in \{1,\ldots,N\}$}, and $f$ and $c$ are nonlinear functions that
we define shortly.  Next, the context vector $z_t$ (computed by the
aligner) encodes the language instruction at time $t \in \{1,\ldots,T\}$.
Next, another RNN decodes the context vector $z_t$ to arrive at the desired
likelihood~\eqref{eqn:argmax}
\begin{subequations} \label{eqn:cond_prob}
    \begin{align}
      P(a_{1:T}|y_{1:T}, x_{1:N}) &= \prod\limits_{t=1}^{T} P(a_t \vert
                   a_{1:t-1}, y_t, x_{1:N})\\
      P(a_t \vert a_{1:t-1}, y_t, x_{1:N}) &= g(s_{t-1},z_t,y_t),
    \end{align}
\end{subequations}
where $s_{t-1}$ is the decoder hidden state at time $t-1$, and $g$ is a
nonlinear function.  Inference then follows by maximizing this posterior to
determine the desired action sequence.

Our model (Fig.~\ref{fig:model}) employs LSTMs as the nonlinear functions
$f$ and $g$ due to their ability to learn long-term dependencies that exist
over the instruction and action sequences, without suffering from exploding
or vanishing gradients. Our model also integrates \emph{multi-level}
alignment to focus on parts of the instruction that are more salient to the
current action at multiple levels of abstraction. We next describe each
component of our network in detail.
\begin{figure}[!t]
    \centering
    \includegraphics[width=1.0\linewidth]{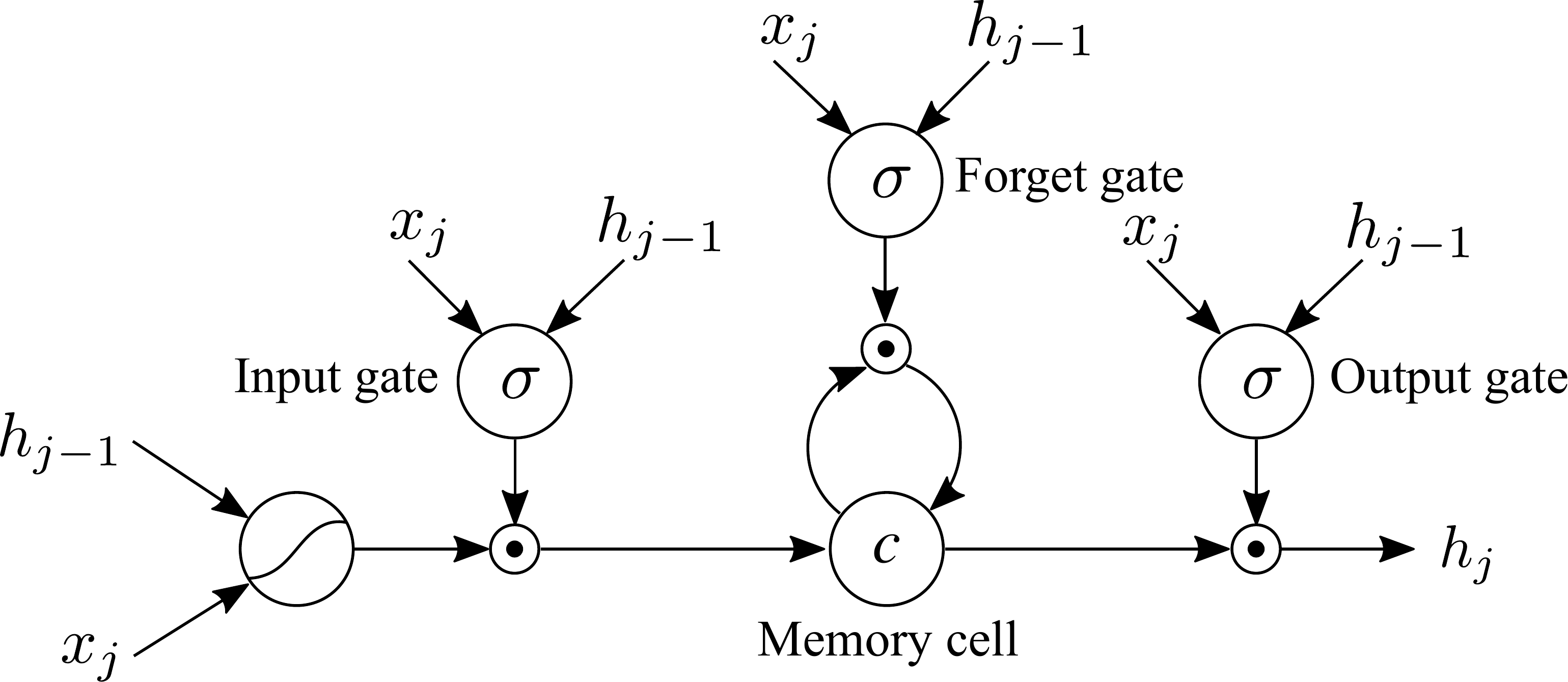}
    \caption{Long Short-term Memory (LSTM) unit.} \label{fig:lstm}
\end{figure}

\paragraph{Encoder} Our encoder takes as input the natural language route
instruction represented as a sequence
\mbox{$x_{1:N} = (x_1, x_2, \ldots, x_N)$}, where $x_1$ and $x_N$ are the
first and last words in the sentence, respectively. We treat each word
$x_i$ as a $K$-dimensional one-hot vector, where $K$ is the vocabulary
size. We feed this sequence into an LSTM-RNN that summarizes the temporal
relationships between previous words and returns a sequence of hidden
annotations $h_{1:N} = (h_1,h_2,\ldots,h_N)$, where the annotation $h_j$
summarizes the words up to and including $x_j$.

We adopt an LSTM encoder architecture
(Fig.~\ref{fig:lstm}) similar to that
of~\myciteauthoryear{graves-13},
\begin{subequations} \label{eqn:encoder}
    \begin{align}
      \begin{pmatrix}
          i^{e}_j\\ 
          f^{e}_j\\
          o^{e}_j\\ 
          g^{e}_j 
      \end{pmatrix} 
   &= 
     \begin{pmatrix}
         \sigma\\
         \sigma\\
         \sigma\\ 
         \tanh 
     \end{pmatrix}
      T^{e} 
      \begin{pmatrix}
          x_j\\ 
          h_{j-1}
      \end{pmatrix}\\
      c^{e}_j&=f^{e}_j \odot c^{e}_{j-1} + i^{e}_j \odot g^{e}_j \\
      h_j &=o^{e}_j \odot \tanh(c^{e}_j) \label{eqn:encoder_h}
    \end{align}
\end{subequations}
where $T^{e}$ is an affine transformation, $\sigma$ is the logistic sigmoid
that restricts its input to $[0,1]$, $i^{e}_j$, $f^{e}_j$, and $o^{e}_j$
are the input, output, and forget gates of the LSTM, respectively, and
$c^e_j$ is the memory cell activation vector. The memory cell $c^e_j$
summarizes the LSTM's previous memory $c^e_{j-1}$ and the current input,
which are modulated by the forget and input gates, respectively. The forget
and input gates enable the LSTM to regulate the extent to which it forgets
its previous memory and the input, while the output gate regulates the
degree to which the memory affects the hidden state.

Our encoder employs bidirectionality, encoding the sentences in both the
forward and backward directions, an approach that has been found to be
successful in speech recognition and machine
translation~\cite{graves-13,bahdanau-14,cho14}. In this way, the hidden
annotations
\mbox{$h_j = (\overrightarrow{h}_j^\top; \overleftarrow{h}_j^\top)^\top$}
concatenate forward $\overrightarrow{h}_j$ and backward annotations
$\overleftarrow{h}_j$, each determined using Equation~\ref{eqn:encoder_h}.

\paragraph{Multi-level Aligner} The context representation of the
instruction is computed as a weighted sum of the word vectors $x_j$ and
encoder states $h_j$. Whereas most previous work align based only on the
hidden annotations $h_j$, we found that also including the original input
word $x_j$ in the aligner improves performance.  This multi-level
representation allows the decoder to not just reason over the high-level,
context-based representation of the input sentence $h_j$, but to also
consider the original low-level word representation $x_j$. By adding $x_j$,
the model offsets information that is lost in the high-level abstraction of
the instruction. Intuitively, the model is able to better match the salient
words in the input sentence (e.g., ``easel'') directly to the corresponding
landmarks in the current world state $y_t$ used in the decoder. The context
vector then takes the form
\begin{equation} \label{eqn:multi-level-aligner}
    z_t = \sum_{j}\alpha_{tj} 
    \begin{pmatrix}
        x_j \\
        h_j
    \end{pmatrix}
\end{equation}
The weight $\alpha_{tj}$ associated with each pair $(x_j, h_j)$ is
\begin{equation}
    \alpha_{tj} = \exp(\beta_{tj})/\sum_j \exp(\beta_{tj}),
\end{equation}
where the alignment term $\beta_{tj}=f(s_{t-1},x_j,h_j)$ weighs the extent
to which the word at position $j$ and those around it match the output at
time $t$. The alignment is modelled as a one-layer neural perceptron
\begin{align}
  \beta_{tj} = v^\top\tanh(Ws_{t-1}+Ux_{j}+Vh_{j}),
\end{align}
where $v$, $W$, $U$, and $V$ are learned parameters.

\paragraph{Decoder} Our architecture uses an LSTM decoder
(Fig.~\ref{fig:lstm}) that takes as input the current world state $y_t$,
the context of the instruction $z_t$, and the LSTM's previous hidden state
$s_{t-1}$. The output is the conditional probability distribution
\mbox{$P_{a,t} = P(a_t \vert a_{1:t-1}, y_t, x_{1:N})$} over the next
action \eqref{eqn:cond_prob}, represented as a deep output
layer~\cite{pascanu-14}
\begin{subequations}
    \begin{align}
      \begin{pmatrix}
          i^d_t\\
          f^d_t\\
          o^d_t\\
          g^d_t
      \end{pmatrix}
   &= 
     \begin{pmatrix}
         \sigma\\ 
         \sigma\\ 
         \sigma\\ 
         \tanh 
     \end{pmatrix} T^d 
      \begin{pmatrix}
          Ey_t\\ 
          s_{t-1}\\ 
          z_t
      \end{pmatrix}\\
      c^d_t&=f^d_t \odot c^d_{t-1} + i^d_t \odot g^d_t \\
      s_{t}&=o_t^d \odot \tanh(c^d_t)\\
      q_t &= L_0 ( Ey_t + L_ss_t + L_zz_t )\\
      P_{a,t} &= \textrm{softmax}\left(q_t\right)
    \end{align}
\end{subequations}
where $E$ is an embedding matrix and $L_0$, $L_s$, and $L_z$ are parameters
to be learned.

\paragraph{Training} We train the encoder and decoder models so as to
predict the action sequence $a^*_{1:T}$ according to
Equation~\ref{eqn:argmax} for a given instruction $x_{1:N}$ and world state
$y_{1:T}$ from the training corpora. We use the negative log-likelihood of
the demonstrated action at each time step $t$ as our loss function,
\begin{align}
  L = -\log P(a_t^*|y_t, x_{1:N}).
\end{align}
As the entire model is a differentiable function, the parameters can
be learned by back-propagation. 

\paragraph{Inference} Having trained the model, we generate action
sequences by finding the maximum a posteriori actions under the learned
model~\eqref{eqn:argmax}.  One action sequence is completed when the
``stop'' action is emitted.  For the single-sentence task, we perform this
search using standard beam search to maintain a list of the current $k$
best hypotheses.\footnote{We use a beam width of 10 to be consistent with
  existing work.} We iteratively consider the $k$-best sequences up to time
$t$ as candidates to generate sequences of size $t+1$ and keep only the
resulting best $k$ of them.  For multi-sentence, we perform the search
sentence-by-sentence, and initialize the beam of the next sentence with the
list of previous $k$ best hypotheses. As a common denoising method in deep
learning~\cite{sutskever-14,zaremba-14,vinyals-14}, we perform inference
over an ensemble of randomly initialized models.\footnote{At each time step
  $t$, we generate actions using the avg.\ of the posterior likelihoods of
  10 ensemble models, as in previous work.}

\section{Experimental Setup}
\label{sec:experimentalsetup}
	
\paragraph{Dataset} We train and evaluate our model using the publicly
available SAIL route instruction dataset collected
by~\myciteauthoryear{macmahon-06}.  We use the raw data in its
original form (e.g., we do not correct any spelling errors).  The
dataset contains $706$ non-trivial navigational instruction
paragraphs, produced by six instructors for $126$ unique start and end
position pairs spread evenly across three virtual worlds. These
instructions are segmented into individual sentences and paired with
an action sequence. See corpus statistics in
\myciteauthoryear{chen-11}.

\paragraph{World State} The world state $y_t$ encodes the local,
observable world at time $t$. We make the standard assumption that the
agent is able to observe all elements of the environment that are
within line-of-sight. In the specific domain that we consider for
evaluation, these elements include the floor patterns, wall paintings,
and objects that are not occluded by walls. We represent the world
state as a concatenation of a simple bag-of-words vector for each
direction (forward, left, and right). The choice of bag-of-words
representation avoids manual domain-specific feature-engineering and
the combinatoriality of modeling exact world configurations.

\paragraph{Evaluation Metrics} We evaluate our end-to-end model on
both the single-sentence and multi-sentence versions of the
corpus. For the single-sentence task, following previous work, the
strict evaluation metric deems a trial to be successful iff the final
position and orientation exactly match those of the original
demonstration. For multi-sentence, we disregard the final orientation,
as previous work does. However, this setting is still more challenging
than single-sentence due to cascading errors over individual
sentences.

\paragraph{Training Details} We follow the same procedure as
\myciteauthoryear{chen-11}, training with the segmented data and
testing on both single- and multi-sentence versions. We train our
models using three-fold cross-validation based on the three maps. In
each fold, we retain one map as test and partition the two-map
training data into training ($90\%$) and validation ($10\%$) sets, the
latter of which is used to tune hyperparameters.\footnote{We only
  tuned the no.\ of hidden units and the drop-out
  rate~\cite{srivastava-14,zaremba-14}.} We repeat this process for
each of the three folds and report (size-weighted) average test
results over these folds. We later refer to this training procedure as
``vDev.'' Additionally, some previous methods (p.c.) adopted a
slightly different training strategy whereby each fold trains on two
maps and uses the test map to decide on the stopping iteration. In
order to compare against these methods, we also train a separate
version of our model in this way, which we refer to as ``vTest.''

For optimization, we found Adam~\cite{kingma-15} to be very effective
for training with this dataset. The training usually converges within
50 epochs. We performed early stopping based on the validation task
metric. Similar to previous work~\cite{xu-15}, we found that the
validation log-likelihood is not well correlated with the task metric.

\section{Results and Analysis}
\label{sec:results}

In this section, we compare the overall performance of our model on
the single- and multi-sentence benchmarks against previous
work. We then present an
analysis of our model through a series of ablation studies.
\begin{table}[!ht]
    \centering
    \caption{Overall accuracy (state-of-the-art in bold)}\label{result}
    \begin{tabularx}{1.0\linewidth}{l c c}
        \toprule
		Method & Single-sent & Multi-sent \\
        \midrule
		\myciteauthoryear{chen-11} & $54.40$ & $16.18$ \\
		\myciteauthoryear{chen-12} & $57.28$ & $19.18$ \\
		\myciteauthoryear{kim-12} & $57.22$ & $20.17$ \\
		\myciteauthoryear{kim-13} & $62.81$ & $26.57$ \\
		\myciteauthoryear{artzi-13} & $65.28$ & $31.93$ \\
		\myciteauthoryear{artzi-14} & $64.36$ & $\mathbf{35.44}$ \\
		\myciteauthoryear{andreas-15} & $59.60$ & -- \\
        Our model (vDev) & $69.98$ & $26.07$ \\
        Our model (vTest) & $\mathbf{71.05}$ & $30.34$ \\
        \bottomrule
    \end{tabularx}
\end{table}
\begin{table*}[!ht]
    \centering
    \caption{Model components ablations}\label{ablation}
    \begin{tabularx}{0.80\linewidth}{l c c c c c}
      \toprule
      & Full Model & High-level Aligner & No Aligner & Unidirectional & No Encoder \\
      \midrule
      Single-sentence & $69.98$  & $68.09$ & $68.05$ & $67.44$ & $61.63$ \\
      Multi-sentence & $26.07$ &  $24.79$ & $25.04$ & $24.50$ & $16.67$ \\
      \bottomrule
    \end{tabularx}
\end{table*}
\paragraph{Primary Result} We first investigate the ability to
navigate to the intended destination for a given natural language
instruction. Figure~\ref{fig:mapexample} illustrates an output example
for which our model successfully executes the input natural language
instruction. Table~\ref{result} reports the overall accuracy of our
model for both the single- and multi-sentence settings. We report two
statistics with our model (vDev and vTest) in order to directly
compare with existing work.\footnote{Subsequent evaluations are on
  vDev unless otherwise noted.}

As we can see from Table~\ref{result}, we surpass state-of-the-art
results on the single-sentence route instruction task (for both vDev
and vTest settings), despite using no linguistic knowledge or
resources. Our multi-sentence accuracy, which is working with a really
small amount of training data (a few hundred paragraph pairs), is
competitive with state-of-the-art and outperforms several previous
methods that use additional, specialized resources in the form of
semantic parsers, logical-form lexicons, and re-rankers.\footnote{Note
  that with no ensemble, we are still state-of-the-art on
  single-sentence and better than all comparable approaches on
  multi-sentence: Artzi et al.\ (2013, 2014) use extra annotations with
  a logical-form lexicon and \myciteauthoryear{kim-13} use
  discriminative reranking, techniques that are orthogonal to our
  approach and should likely improve our results as well.}  We note
that our model yields good results using only greedy search
(beam width of one). For vDev, we achieve $68.05$ on single-sentence
and $23.93$ on multi-sentence, while for vTest, we get $70.56$ on
single-sentence and $27.91$ on multi-sentence.

\paragraph{Distance Evaluation} Our evaluation required that the
action sequence reach the exact desired destination. It is of interest
to consider how close the model gets to the destination when it is not
reached. Table~\ref{close} displays the fraction of test results that
reach within $d$ nodes of the destination. Often, the method produces
action sequences that reach points close to the desired destination.
\begin{table}[!ht]
    \centering
    \caption{Accuracy as a function of distance from destination} \label{close}
    \begin{tabularx}{0.9\linewidth}{lccccccc}
        \toprule
        Distance ($d$) & $0$ & $1$ & $2$ & $3$\\
        \midrule
        Single-sentence & $71.73$ & $86.62$ & $92.86$ & $95.74$ \\ 
        Multi-sentence & $26.07$ & $42.88$ & $59.54$ & $72.08$ \\ 
        \bottomrule
    \end{tabularx}
\end{table}

\begin{figure*}[th]
    \centering
    \includegraphics[width=0.95\linewidth]{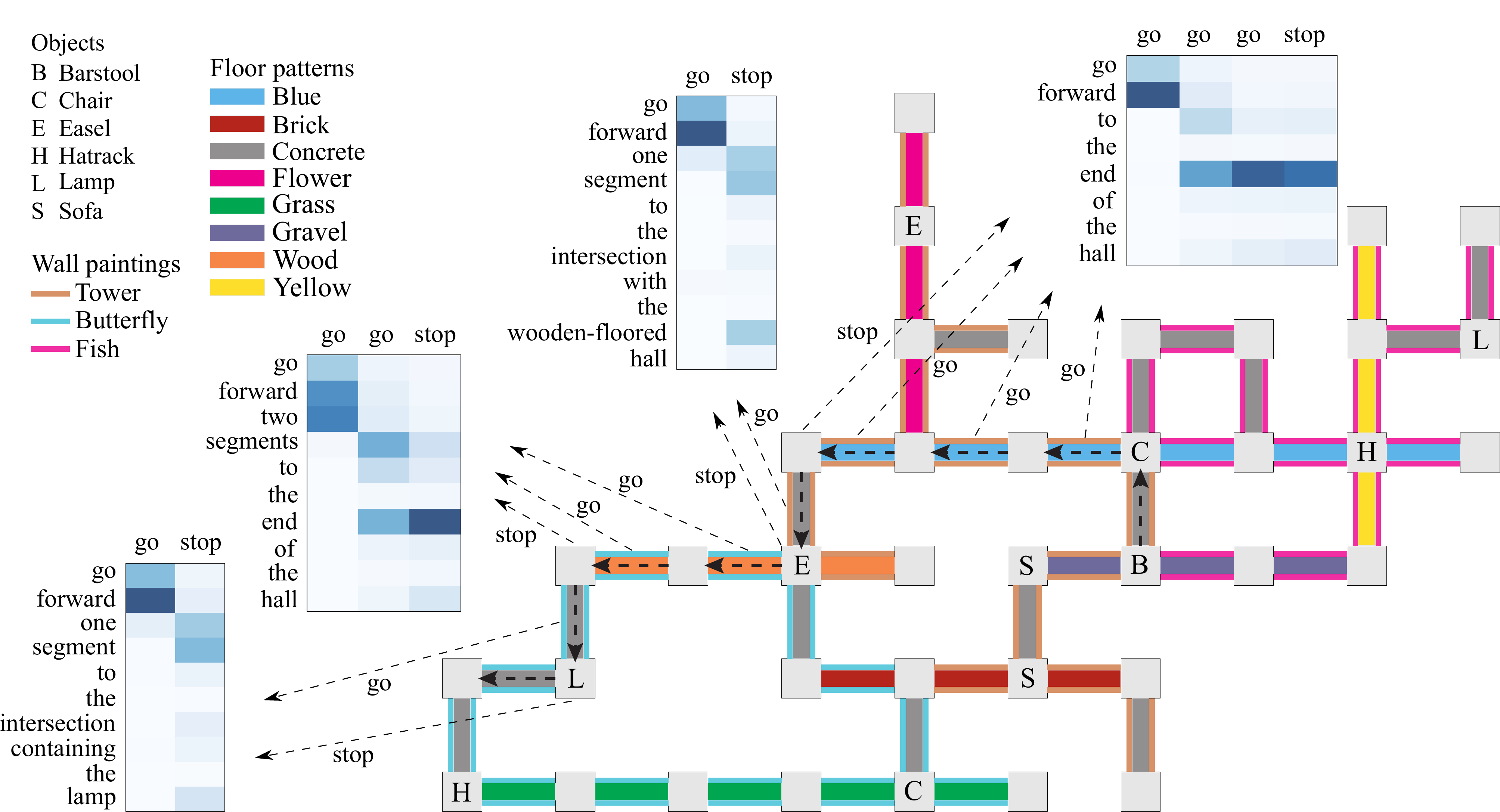}
    \caption{Visualization of the alignment between words to actions
      in a map for a multi-sentence instruction.} \label{fig:alignexample}
\end{figure*}

\paragraph{Multi-level Aligner Ablation} 
Unlike most existing methods that align based only on the hidden
annotations $h_j$, we adopt a different approach by also including the
original input word $x_j$ (\eqnref{eqn:multi-level-aligner}). As shown
in Table~\ref{ablation}, the multi-level representation (``Full
Model'') significantly improves performance over a standard aligner
(``High-level Aligner'').  Figure~\ref{fig:alignexample} visualizes
the alignment of words to actions in the map environment for several
sentences from the instruction paragraph depicted in
Figure~\ref{fig:mapexample}.
\paragraph{Aligner Ablation} 
Our model utilizes alignment in the decoder as a means of focusing on
word ``regions'' that are more salient to the current world state. We
analyze the effect of the learned alignment by training an alternative
model in which the context vector $z_t$ is an unweighted average
(Eqn.~\ref{eqn:multi-level-aligner}). As shown in the Table
\ref{ablation}, learning the alignment does improve the accuracy of
the resulting action sequence. Note that the ``No Aligner'' model
still maintains all connections between the instruction and actions,
but is just using non-learned, uniform weights.

\paragraph{Bidirectionality Ablation} We train an alternative
model that uses only a unidirectional (forward) encoder. As shown in
Table~\ref{ablation}, the bidirectional encoder (``Full Model'')
significantly improves accuracy.

\paragraph{Encoder Ablation} We further evaluate the benefit of
encoding the input sentence and consider an alternative model that
directly feeds word vectors as randomly initialized embeddings into
the decoder and relies on the alignment model to choose the salient
words. Table~\ref{ablation} presents the results with and without the
encoder and demonstrates that there is a substantial gain in encoding
the input sentence into its context representation. We believe the
difference is due to the RNN's ability to incorporate sentence-level
information into the word's representation as it processes the
sentence sequentially (in both directions). This advantage helps
resolve ambiguities, such as ``turn right before \ldots'' versus
``turn right after \ldots''.
\section{Conclusion}
\label{sec:conclusion}

We presented an end-to-end, sequence-to-sequence approach to mapping
natural language navigational instructions to action plans given the
local, observable world state, using a bidirectional LSTM-RNN model with a multi-level
aligner. We evaluated our model on a benchmark route instruction
dataset and demonstrates that it achieves a new state-of-the-art on
single-sentence execution and yields competitive results on the more
challenging multi-sentence domain, despite working with very small
training datasets and using no specialized linguistic knowledge or
resources. We further performed a number of ablation studies to
elucidate the contributions of our primary model components.  
\section*{Acknowledgments}

We thank Yoav Artzi, David Chen, Oriol Vinyals, and Kelvin Xu for
their helpful comments. This work was supported in part by the
Robotics Consortium of the U.S. Army Research Laboratory under the
Collaborative Technology Alliance Program, Cooperative Agreement
W911NF-10-2-0016, and by an IBM Faculty Award.

\bibliography{references}
\bibliographystyle{aaai}

\end{document}